\documentclass[conference]{IEEEtran}
\IEEEoverridecommandlockouts
\usepackage{cite}
\usepackage{amsmath,amssymb,amsfonts}
\usepackage{algorithmic}
\usepackage{graphicx}
\usepackage{textcomp}
\usepackage{xcolor}
\usepackage{array}
\usepackage{xcolor}
\def\BibTeX{{\rm B\kern-.05em{\sc i\kern-.025em b}\kern-.08em
    T\kern-.1667em\lower.7ex\hbox{E}\kern-.125emX}}
\begin{document}

\title{Learning Personal Food Preferences via Food Logs Embedding
}


\author{Ahmed A. Metwally$^{1,2,*}$, Ariel K. Leong$^{3,*}$, Aman Desai$^{4}$, Anvith Nagarjuna$^{1}$, Dalia Perelman$^{1}$, Michael Snyder$^{1}$

\thanks{$^{1}$ Department of Genetics, Stanford University, Stanford, CA, USA}
\thanks{$^{2}$ Systems and Biomedical Engineering Department, Faculty of Engineering, Cairo University, Giza, Egypt}
\thanks{$^{3}$ Department of Computer Science, Stanford University, Stanford, CA, USA}
\thanks{$^{4}$ Department of Computer Science, University of California at Berkeley, Berkeley, USA}
\thanks{$^{*}$ These authors contributed equally}
\thanks{Correspondence: {\emph{\small ametwall@stanford.edu}}}
\thanks{This work was supported by the Stanford PHIND award.}
}

\maketitle

\begin{abstract}
Diet management is key to managing chronic diseases such as diabetes. Automated food recommender systems may be able to assist by providing meal recommendations that conform to a user's nutrition goals and food preferences. Current recommendation systems suffer from a lack of accuracy that is in part due to a lack of knowledge of food preferences. In this work, we propose a method for learning food preferences from food logs, a comprehensive but noisy source of information about users' dietary habits. We also introduce accompanying metrics to evaluate personal learning food preferences. The method generates and compares word embeddings to identify the parent food category of each food entry and then calculates the most popular. Our proposed approach identifies 82\% of a user's ten most frequently eaten foods. Our method is publicly available on (https://github.com/aametwally/LearningFoodPreferences). 
\end{abstract}

\begin{IEEEkeywords}
Food Preferences, Recommender Systems, Machine Learning.
\end{IEEEkeywords}

\section{Introduction}
Millions of Americans suffer from chronic diseases whose management could be greatly influenced by diet. Following a dietary recommendation is more successful when the patient or consumer is familiar with the food and its preparation. Automated food recommendation systems can assist patients in identifying healthy meals that align with their diet preferences. However, current food recommendation systems struggle to achieve high accuracies \cite{trattner2017food}. One reason is a lack of focus on learning users' food preferences. They do not focus on what foods users actually eat. Many food recommendation systems use classic strategies such as content-based filtering (CB) and collaborative filtering (CF). The former involves recommending items similar to ones that the user likes. The latter builds on this approach by recommending items that are similar to both ones the user has previously liked and to ones liked by similar users. Examples of CB- and CF- approaches include that used by Forbes and Zhu \cite{forbes2011content}, who used a collaborative filtering algorithm but improved it by directly incorporating information about a recipe's ingredients into the matrix factorization part of the algorithm. Another group created an equation to predict the score a user would assign to a recipe that took into account the number of ingredients in the recipe that the user liked and how much they liked them \cite{harvey2013you}. There are several disadvantages to these approaches. Their effectiveness relies on having detailed information about users' feelings, specifically the user for whom the recommendation is made and similar users, toward many different items. Such detailed information is difficult to obtain.
Additionally, many of the previous studies either use data scraped from recipe rating websites or present users with a series of recipes and ask for their rating \cite{trattner2017food}. This data is likely not fully representative of users' eating patterns, as most users do not input ratings for all foods they eat. They probably only review foods they feel especially strongly (positively or negatively) about. Additionally, many of these methods do not consider the frequency with which a user eats a dish. This is important information that influences how likely a user would be to eat a recommended dish.

In addition to CB- and CF-based methods, others have taken innovative approaches to gauge food preferences in their recommender systems. Ueda et al. use a user's food log to gauge how much users like various recipe ingredients based on their frequency of use \cite{ueda2011user}. Another group successively presents images to a user, using a convolutional neural network (CNN) algorithm to learn a user's preferences \cite{yang2017yum}. While the approach is innovative and interesting, it is unclear whether people base eating decisions on food appearance when preparing meals at home (instead of ordering in a restaurant). Toledo et al. integrate food preferences into their recommendation approach by devising menus containing ingredients that users have liked in the past but not eaten recently and revising based on user feedback of which ingredients they like and do not like \cite{toledo2019food}. Other studies treat food recommendations as a query over a knowledge graph. One takes in a food log or list of liked foods and allergies and outputs recipes that are most similar to the input \cite{chen2021personalized}. Another outputs sets of foods that are predicted to pair well together \cite{park2021flavorgraph}. However, these predictions do not take in user input.

Food logs have been used for purposes such as allergy detection and weight and disease management \cite{cordeiro2015barriers}. One common difficulty has been consistently recording entries for a variety of reasons, such as the laboriousness involved in recording certain kinds of foods, negative emotions arising when journaling and lack of social support \cite{ye2016assisting}. Recent food logging applications such as MyFitnessPal \cite{myfitnesspal}, Cronometer \cite{cronometer}, and Lose It! \cite{loseit} have addressed these challenges by providing a platform that allows the user to input the food name, the quantity, the type of meal, and the time at which the user consumed the food. There can be gamification features or social support to address the barriers to journaling mentioned earlier. However, there are several shortcomings in using the food logs exported directly from these food logging apps. Food names can contain specific brand names, and the food log structure can differ from one website to another, making them difficult to streamline data processing and information retrieval. Our proposed approach handles all of these inconsistencies.

Word embeddings have been deployed in food computing because they can capture relationships between different ingredients and concepts. Recipe embeddings have been learned, for example, to build ingredient maps \cite{morales2021word}. Diet2Vec used food name embeddings to model meals, and diets \cite{tansey2016diet2vec}. They used data from Lose It! journals. They showed that similar foods could cluster together from embeddings that were 20\% name and 80\% nutrition. They also averaged words to create food name vectors.

In this work, we propose a method to identify food preferences from food logs that also includes: 1) creating a method to gauge food preferences that uses food logs, 2) using a publicly available dataset, the U.S. Department of Agriculture's Food and Nutrient Database for Dietary Studies \cite{FNDDS}, 3) using evaluation metrics that others can easily adopt, and 4) making our code publicly available. Our system identifies food preferences by identifying frequently eaten foods. The method first involves assigning each food log entry one of the labels used in the U.S. Department of Agriculture's Food and Nutrient Database for Dietary Studies (FNDDS).

\section{Methods}

\begin{figure*}[!ht]
\centering
\includegraphics[width=\textwidth]{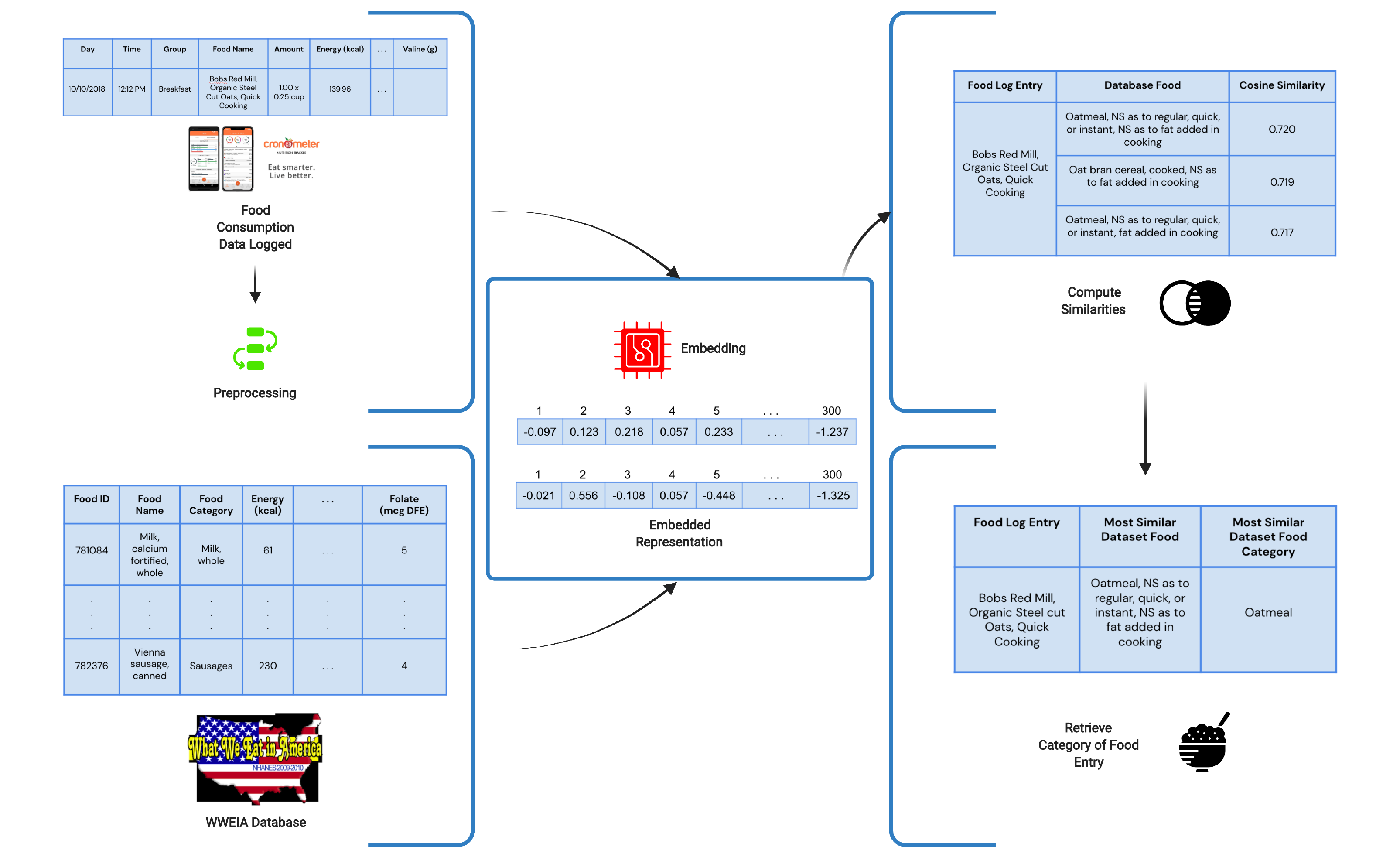}
\caption[]{Workflow of the food preference learning algorithm. Each entry of the food log is processed through an NLP module. Then food embeddings is obtained for each food entry in the food log and the the database. Next, for each food embedding of the food log, Cosine similarity is run against all embeddings of the foods in the WWEIA database. The food category label of the food with the highest Cosine similarity is assigned to the food log entry. This process is repeated for all foods in the food logs. The most common food categories are then calculated.}
\label{MethodFlowchart}
\end{figure*}

The method, as detailed in Figure~\ref{MethodFlowchart}, has 4 components: preprocessing food logs, generating embedding vectors for food log and database entries, computing vector similarities, and using these similarities to identify commonly eaten foods. Evaluation metrics were also developed.

\subsection{Preprocessing of Food Logs and Database}

\subsubsection{Food Logs}
Food logs were obtained from Cronometer. This food tracking app was chosen because its nutrition database is company-maintained instead of allowing user contributions. (Accurate, comprehensive nutrition information would be helpful for future food recommendations.) A sample of a food log is shown in Figure~\ref{MethodFlowchart}. Visualizations of food log statistics can be found in Figure~\ref{FoodlogsSummary} and Figure~\ref{fooddistribution}. Each entry in a food log was labeled with the date, time, and description of the food eaten. Food names consisted of a series of comma-separated phrases. An example is "Trader Joe's, Chicken Sausage, Sweet Italian Style." The first phrase contained either the food manufacturer's brand name, the brand name, the food name or only the food name. The following phrases in the food name entry contained specific details about food composition or preparation.

\subsubsection{Database}
The database used was the U.S. Department of Agriculture's Food and Nutrient Database for Dietary Studies (FNDDS) database. A sample of the database is shown in Figure~\ref{MethodFlowchart}. It consists of 8691 foods belonging to 155 categories that were recorded during the dietary intake component of the National Health and Nutrition Examination Survey. The FNDDS was chosen because it is representative of foods eaten in America across demographics, contains detailed nutritional information that will be useful for the future food recommendation component, and already includes annotations mapping each food to a category. Each category is a different type of food; examples of categories can be found in Figure~\ref{fooddistribution}. The number of foods belonging to a category varies widely; the mean number is 56.06, and the standard deviation is 69.97. There are 19 categories with fewer than ten foods. FNDDS food names consist of a series of comma-separated phrases. An example is "Yogurt, Greek, whole milk, fruit."

The descriptive name and assigned label of each food entry in the FNDDS database were extracted. Foods that were assigned a label pertaining to baby foods or formula were removed. Food log processing consisted of extracting food log entry names and annotating them with the list of food labels used in the FNDDS.

\subsection{Learning Food Preferences}
Our proposed system identifies food preferences by identifying foods that are eaten frequently. The system first labels each food log entry with one of the 155 food labels used in the Food and Nutrient Database for Dietary Studies (FNDDS). The ten most popular foods are then calculated. 

\subsubsection{Labeling with Food Embeddings}
Due to the small dataset size, a relatively large number of classification categories, and a small number of samples in each category, k-Nearest Neighbors Classification with $k = 1$ was used. Food numerical representation (of length 300) for all food log entries as well as all FNDDS entries were generated using Word2Vec, which is a word-embedding model word pretrained on the Google News database. We finetuned Word2Vec on artificial sentences that concatenated the FNDDS' descriptive food names and the category the food belonged to \cite{mikolov2013efficient}. Embeddings for food names, which consist of multiple words, are formed by averaging the embeddings of the constituent words. The cosine similarity between the embedding for each food log entry and each entry in the database was calculated:

\begin{equation}
    Cosine(x,y) = \frac{x \cdot y}{|| x || || y ||}
\end{equation}

where $x$ is the embedding for a food log entry and $U$ contains the embeddings for all entries in the database. The label of the database food with the highest similarity was used as the system's prediction for the food log entry's label.

\begin{equation}
    Label(x) = argmax_{y \in U} Cosine(x,y) 
\end{equation}

\subsubsection{Food Log Name Preprocessing Methods}

\begin{table}[!ht]
\centering
\label{FoodProcessExample}
\caption{Food Log Name Preprocessing Methods Examples}
\resizebox{0.5\textwidth}{!}{
\begin{tabular}{c p{5.0cm}}
\hline
\textbf{Method}       &
\textbf{Result of Processing the Phrase "Panera Bread, salad, cobb, green goddess, with chicken \& dressing"} \\ \hline
1 & "bread" \\
2 & "bread, salad, cobb, green, with, chicken, \&, dressing" $\xrightarrow{}$ embedding \\
3 & "salad, cobb, green, with, chicken, \&, dressing" $\xrightarrow{}$ embedding \\
4 & "salad, cobb, green, chicken, dressing" \\
5 & "bread, salad, cobb, green, chicken, dressing" $\xrightarrow{}$ embedding \\
6 & "salad, cobb, green, chicken, dressing" but the database is restricted to foods in the "Vegetable mixed dishes", "Chicken, whole pieces", "Chicken patties, nuggets, and tenders", "Yeast breads," and "Salad dressings and vegetable oils" categories $\xrightarrow{}$ embedding  \\
\end{tabular}
}
\end{table}

In our approach, the correct labeling of a food depends on the similarity between a food log entry and its database counterpart (or the one most closely analogous). Thus to improve food label accuracy, food log entry names were preprocessed in various ways to remove words that increased similarity to incorrect FNDDS entries or decreased similarity to the correct entry. An example showing the preprocessing strategies applied to one food log entry, "Panera Bread, salad, cobb, green goddess, with chicken \& dressing," is shown in Table~\ref{FoodProcessExample}. The preprocessing methods, which build on each other, are described below:

\begin{itemize}
    \item Method 1 was intended to only retain the food's general name. As previously mentioned, Cronometer food log entry names were structured as a series of comma-separated phrases. Generally, the first phrase would contain the general food name unless the food brand was specified. The food brand would be in the first comma-separated phrase, and the general food would be the second comma-separated phrase. Additional phrases specifying further details about the food would follow it. The first phrase would contain the brand name if there was one, followed by the food name. The remaining phrases contained food composition or preparation details. Removing the food manufacturer's brand name and more specific details of the foods was hypothesized to increase the similarity between analogous foods. The removal was accomplished by eliminating all but either the first or second comma-separated phrase before generating an embedding.
    Determining whether the first comma-separated phrase contained a food brand name and should not be used consisted of counting the number of words in the first and second comma-separated phrases that belonged to the FNDDS vocabulary and choosing the phrase that had the larger number.
    \item Method 2 was similar to Method 1, but retained the comma-separated phrases that contained specific details.
    \item Method 3, like Method 2, retained most of the food name, but used another heuristic to judge whether the first comma-separated phrase contained a brand name. Instead of counting the number of words that belonged to the FNDDS vocabulary, the percentage of FNDDS words in the comma-separated phrase was used. \item Method 4, in which generic food-related terms were removed from the food log entry name (in addition to the preprocessing done in Method 3), was introduced after noticing that for some food log entries, the most similar database food was one that was wholly unrelated but contained a generic word in common. For example, the most similar database food for many fruits, including "Blueberries, Fresh," "Blackberries, Fresh," and "Strawberries, Fresh" was "Fresh corn custard, Puerto Rican style." Thus the frequency of each word in the FNDDS vocabulary was tabulated. All of the generic words among the top 250 most common words were removed from the food log name. 
    \item Method 5 addressed the mislabeling of foods such as "Kind, Nuts \& Spices Bar, Dark Chocolate Nuts \& Sea Salt," where the first comma-separated phrase contains not only the brand name but also the general food name. This method was identical to Method 4 except that instead of removing the whole first comma-separated phrase, only words not found in the FNDDS vocabulary were removed. 
    \item Method 6 addressed mislabeling errors where the predicted food label was very different from the true label. (For example, "Orowheat, Thin-Sliced Rustic White" was misclassified as "Liquor and cocktails.") Instead of being compared to the embeddings of all FNDDS foods, a food log entry was only compared to foods whose labels had associated FNDDS foods that shared words in common with the food log name.
\end{itemize}

\subsection{Evaluation Metrics}

\subsubsection{Labeling Accuracy Metrics}
Several evaluation metrics were employed to assess how well the system assigned the correct food label to each food entry. All metrics were calculated individually for each food log and then averaged. Accuracy was computed for each food log as:

\begin{equation}
    Accuracy_{label} = \frac{\text{\# of correct assignments of unique foods}}{\text{\# of unique food log entries}}
\end{equation}

Due to the presence of overlapping food categories (for example, "Milk, whole" and "Milk shakes and other dairy drinks"), the "synonymous accuracy" was also calculated: 

\begin{equation}
    Accuracy_{syn} = \frac{{\text{\# of synonymous assignments of unique foods}}}{\text{\# of unique food log entries}}
\end{equation}

Food categories were considered synonymous if they shared at least one word in common. Accounting for synonymous categories reduced the number of database food categories from 155 to 98. Since it could be the case that the model did not predict the correct label but assigned it a high probability, the mean reciprocal rank (MRR) was also assessed:

\begin{equation}
    MRR = \frac{1}{n}\sum_{i=1}^n\frac{1}{rank_i}
\end{equation}

where $n$ is the total number of unique foods in the food log and $rank$ is the rank that the model assigned each correct label for a food. 

Similar to accuracy, the synonymous mean reciprocal rank (SMRR) was calculated:

\begin{equation}
    SMRR = \frac{1}{n}\sum_{i=1}^n\frac{1}{synrank_i}
\end{equation}

where $synrank$ is the rank of the highest-ranking synonymous food category.

\subsubsection{Identifying Food Preferences Metrics}
Effectiveness at identifying food preferences was evaluated in several ways. Again, all metrics were calculated individually for each food log and then averaged. Food preference accuracy was defined as 

\begin{equation}
    Accuracy_{preference} = \frac{1}{|categories|}\sum_{\substack{i \in categories}}{\{p_{i_d} == p_{i_f}\}}
\end{equation}

where the categories are grains, vegetables, proteins, fruits, and dairy, and $p_{i_d}$ is the most popular food for category $i$ in the dataset, and $p_{i_f}$ is the most popular food for category $i$ in the food log.
A corresponding synonymous accuracy (whether the food identified as the most popular was from a synonymous category) was also calculated. To measure food preferences beyond one favorite, the percentage of the user's top ten most commonly eaten foods that the model was able to identify was also calculated. A synonymous percentage was also calculated.

\section{Results and Discussion}
\subsection{Food Logs and Database Summary}

\begin{figure}[!ht]
\centering
\includegraphics[width=0.45\textwidth]{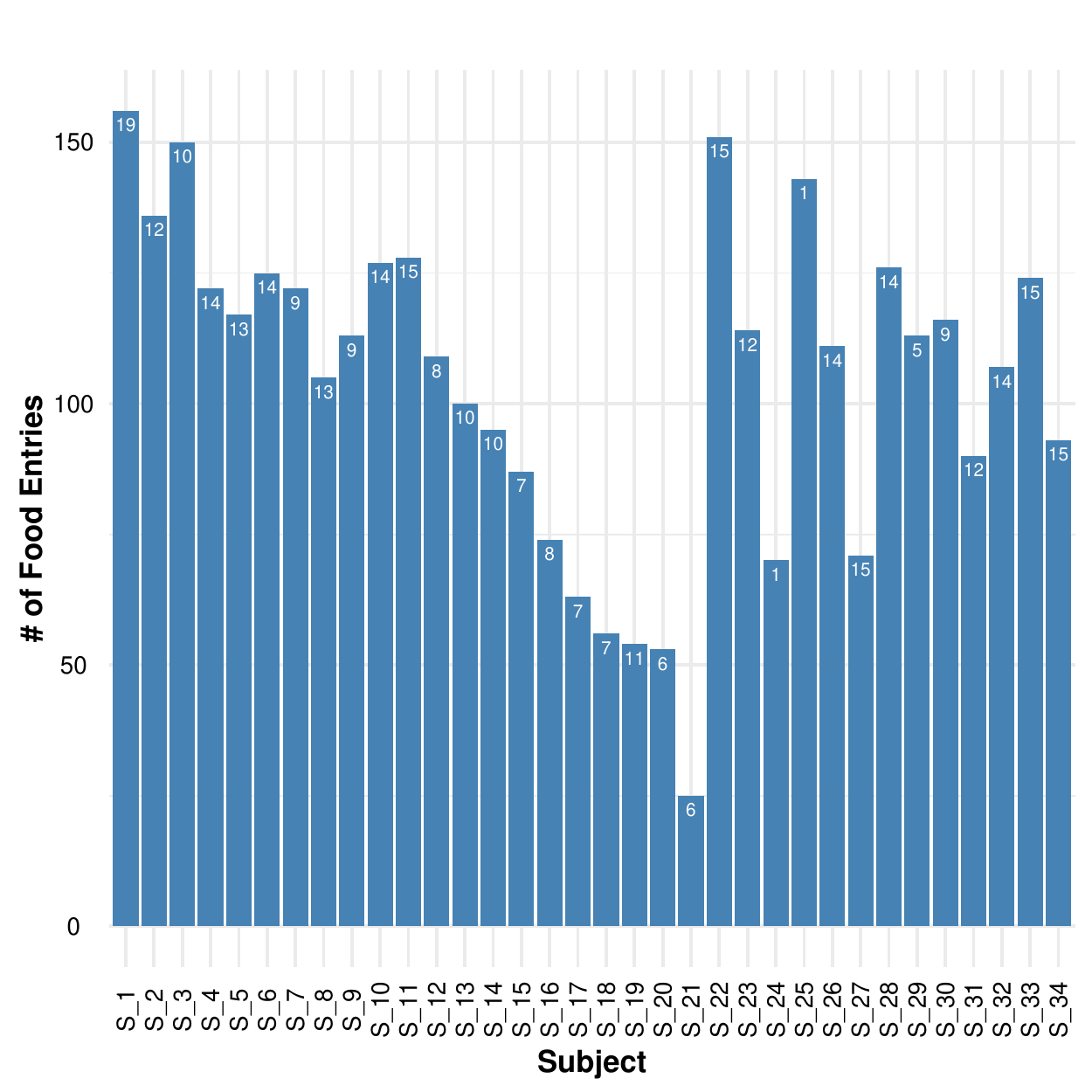}
\caption[Food log]{\textbf{Distribution of number of food entries per food log.} Each bar represents the total number of entries in each subject's food log, with the numbers at the top of each bar displaying the number of days across which the data was being recorded. }
\label{FoodlogsSummary}
\end{figure}

Figure~\ref{FoodlogsSummary} demonstrates that most of the 34 food logs used in the analysis contain over 100 entries. This suggests that the samples contain a representative selection of different foods that can be generalized to a larger population of subjects.

\begin{figure*}[!ht]
\centering
\includegraphics[width=0.9\textwidth]{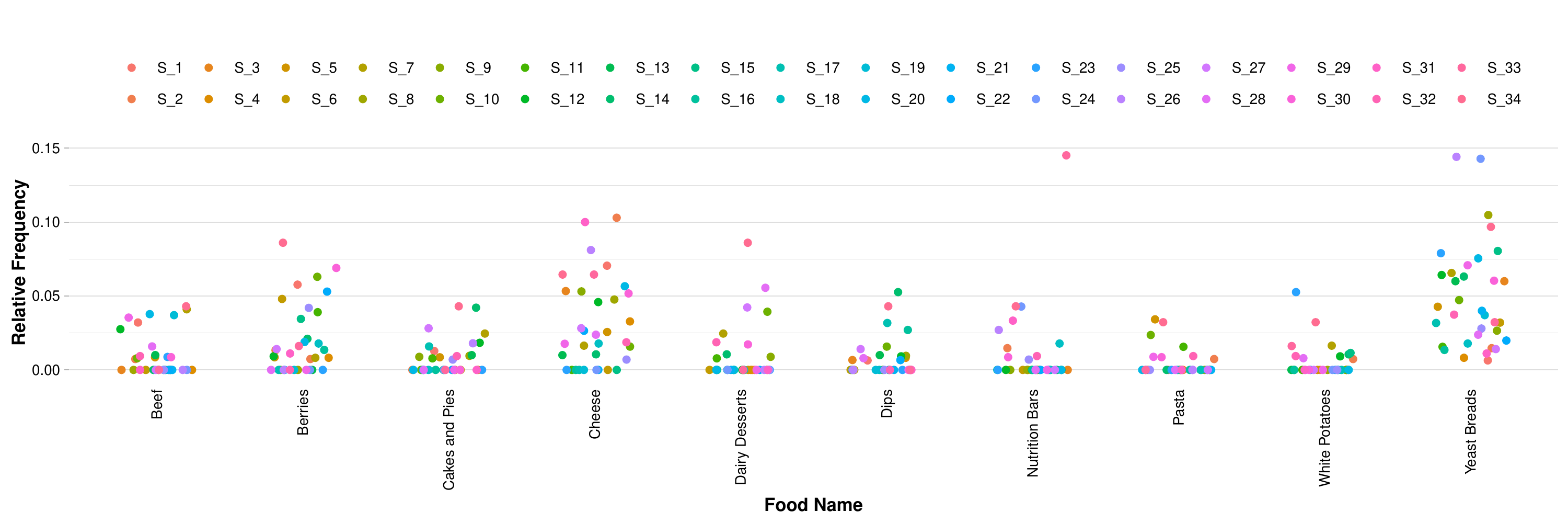}
\caption[]{\textbf{The relative frequency distribution for the ten most common food labels across the food logs.}}
\label{fooddistribution}
\end{figure*}

Figure~\ref{fooddistribution} affirms this idea, illustrating the wide diversity in food selection among the 34 sampled individuals. Even relatively common food item categories like "Yeast breads" rarely make up more than 15\% of any individual's log, although it should be noted that the current food log entry system does not measure the exact amount of food consumed. Interestingly, none of the 10 most chosen food items are meat dishes; this is likely due to the fact that there are a large number of food log categories that refer to dishes with meat in them (ex. "Ground beef", "Pork", "Turkey, duck, other poultry", etc.). Overall, the lack of any overwhelmingly common food category selections indicates that the dataset has enough variation for us to make preliminary conclusions.


\subsection{Performance Evaluation}




\begin{table*}[!ht]
\centering
\label{Evaluation}
\caption{Results of evaluating the food preference learning algorithm on the introduced food labeling and food reference metrics.}
\resizebox{1\textwidth}{!}{
\begin{tabular}{ccccccccc}
\hline
                & \multicolumn{4}{c}{\textbf{Food Labeling}}                                      & \multicolumn{4}{c}{\textbf{Food Preference}}                                                                                            \\ \hline
\textbf{Method} & \textbf{Accuracy} & \textbf{Synonymous Accuracy} & \textbf{MRR} & \textbf{SMRR} & \textbf{Accuracy} & \textbf{Synonymous Accuracy} & \textbf{\% Top 10 Foods Identified} & \textbf{\% Top 10 Synonymous Foods Identified} \\ \hline

1 & 0.42 & 0.48 & 0.49 & 0.55 & 0.41 & 0.47 & 0.46 & 0.76 \\ 
2 & 0.42 & 0.47 & 0.48 & 0.52 & 0.35 & 0.40 & 0.44 & 0.75 \\ 
3 & 0.43 & 0.48 & 0.52 & 0.56 & 0.37 & 0.41 & 0.47 & 0.77 \\ 
\textbf{4} & \textbf{0.49} & \textbf{0.54} & \textbf{0.57} & \textbf{0.62} & \textbf{0.47} & \textbf{0.51} & \textbf{0.52} & \textbf{0.82} \\ 
5 & 0.45 & 0.49 & 0.53 & 0.58 & 0.39 & 0.45 & 0.49 & 0.81 \\ 
6 & 0.30 & 0.37 & \textbf{0.57} & \textbf{0.62} & 0.23 & 0.31 & 0.32 & 0.74 \\

\end{tabular}
}
\end{table*}



Table~\ref{Evaluation} summarize the performance evaluation of each of the 6 methods in terms of food labeling metrics and food preferences metrics. The highest-performing method, Method 4, achieved an accuracy of 49\% and identified 82\% of users' 10 most frequently-eaten foods (with synonymous categories included). A mean reciprocal rank of 0.57 suggests that for many of the food log entries, the correct food label was one of the top two predicted choices. Comparisons to other food preference evaluation work were difficult because their approaches involved different evaluation metrics.

The work shed light on some of the challenges involved in working with food logs. The varying performance of different methods underscored the importance of distinguishing between words that could bias the embeddings and should be removed and words that were instrumental in establishing similarity with the correct analogous database food. Methods 4 and 5's high performance on most of the tasks is likely partly due to the removal of generic food-related words. The poor performance of Method 6, the method that restricted the FNDDS foods that a food log entry was compared with to only those belonging to a category that contained foods that shared at least one word in common with the food log entry, supports the importance of focusing on comparing food log entries based on the contexts of their component words rather than on the words themselves. Method 6 incorrectly labeled foods such as "Creme Fraiche," which was predicted to have the "Doughnuts, sweet rolls, pastries" label since neither "creme" nor "fraiche" appeared in a food name belonging to the correct category, "Cream cheese, sour cream, whipped cream."

Some incorrect label predictions were due to dataset limitations. For several of the non-Western foods in the food logs, such as sev, there were few or no similar foods in the dataset. The database also did not contain alternate spellings, such as "yoghurt," or abbreviations such as "froyo." The heuristic for determining whether the first comma phrase contained a company name was misled by company names that contained food names, such as "Chipotle."

\section{Conclusion \& Future Work}
In this work, we introduce an approach to identify food preferences from food logs that uses embeddings. We also propose accompanying evaluation metrics. Our highest-performing method identifies 82\% of a user's 10 most frequently-eaten foods. This information regarding user's favored foods can be used to generate healthy and realistic meal recommendations that feature ingredients that the user commonly consumes.

Our proposed approach can be generalized to other food logging apps besides Cronometer and other food preference details besides a user's most frequently eaten foods. Each food logging application has its own structure for a food name. This work introduces several methods of preprocessing food log names. For each application, one approach may more accurately identify dietary preferences than another. This method also provides a guide for identifying other dietary preferences using a similar method to the one used to identify frequently-eaten foods: create a set of vectors corresponding to each available option for a dietary preference. For example, if one were trying to identify a user's favored cuisines, they could create vectors for the cuisines "Chinese food" and "Mediterranean food." For each food entry, use cosine similarity to identify the cuisine most likely to belong to that food.


Limitations of the work include the small number of food logs used; annotating all of the entries with the corresponding FNDDS food category required enormous effort. The large decrease in accuracy for Method 6 suggests that using improved embeddings, such as those generated using BERT \cite{devlin2018bert} or ELMo \cite{peters2018deep}, could lead to better performance. The greater incorporation of context into the embeddings could assist in overcoming dataset limitations and labeling food log entries that do not have an identical analog in FNDDS. The increase in performance when common words were removed suggests that other ways of weighting words that differ in importance when determining similarity, such as incorporating the TF-IDF statistic, may lead to improved performance. 

In the future, we plan to add a recommendation component to our food preference learning system. After learning the kinds of foods the user prefers to eat, the system will use nutritional information to recommend healthy variants of the favored foods that fit with the user's metabolic goals. We also plan to evaluate our system with real users and improve the system by taking cuisine preferences into account.

\section{Code Availability}
The project source code is publicly available on (https://github.com/aametwally/LearningFoodPreferences).

\bibliographystyle{unsrt}
\bibliography{FoodPreference}

\end{document}